\documentclass[journal]{IEEEtran}

\usepackage{cite}
\usepackage{enumitem}
\ifCLASSINFOpdf
\usepackage[pdftex]{graphicx}
\usepackage{multicol}
\else
\fi

\usepackage[cmex10]{amsmath}
\usepackage{amssymb}
\usepackage{float}
\usepackage{amsfonts} 
\usepackage{multirow}
\usepackage{array}
\usepackage{tabularx}
\usepackage{tensor}
\usepackage[linesnumbered,ruled,vlined]{algorithm2e}
\usepackage{xcolor}
\usepackage{pifont}
\usepackage{hyperref}
\hypersetup{
    colorlinks=true,
    linkcolor=blue,
    filecolor=magenta,      
    urlcolor=blue
    }
\usepackage{cleveref}

\newcommand{\cmark}{\ding{51}}%
\newcommand{\xmark}{\ding{55}}%

\SetCommentSty{mycommfont}

\SetKwInput{KwInput}{Input}                % Set the Input
\SetKwInput{KwOutput}{Output}              % set the Output

\newcommand\T{{\hspace{-0pt}\intercal}}

\newcolumntype{Z}[0]{>{\hsize=2.4\hsize}X}%
\newcolumntype{s}[0]{>{\centering\arraybackslash\hsize=0.65\hsize}X}%
\newcolumntype{i}[0]{>{\centering\arraybackslash\hsize=1\hsize}X}%

\begin{document}
\title{Self-Reconfigurable Soft-Rigid Mobile Agent with Variable Stiffness and Adaptive Morphology}

\author{Luiza Labazanova,~\IEEEmembership{Student~Member,~IEEE}, Shuang Peng, Liuming Qiu, Hoi-Yin Lee,~\IEEEmembership{Student~Member,~IEEE}\\ Thrishantha Nanayakkara,~\IEEEmembership{Senior Member,~IEEE}, and David Navarro-Alarcon,~\IEEEmembership{Senior Member,~IEEE}% 
\thanks{This work is supported in part by the Research Grants Council (RGC) of Hong Kong under grants 14203917 and 15212721, and in part by the 2019/20 Belt and Road Scholarship (Research Postgraduate).}%
\thanks{L. Labazanova, L. Qiu, H. Lee, and D. Navarro-Alarcon are with the Faculty of Engineering, The Hong Kong Polytechnic University, HK (e-mail: luiza.labazanova,valen.qiu,hoi-yin.lee@connect.polyu.hk, dna@ieee.org).}%
\thanks{S. Peng is with the Department of Aerospace and Mechanical Engineering, University of Southern California, USA. (e-mail: shuangpe@usc.edu).}
\thanks{T. Nanayakkara is with the Dyson School of Design Engineering, Imperial College London, UK (e-mail: t.nanayakkara@imperial.ac.uk).}
}

\maketitle

\begin{abstract}
In this paper, we propose a novel design of a hybrid mobile robot with controllable stiffness and deformable shape. 
Compared to conventional mobile agents, our system can switch between rigid and compliant phases by solidifying or melting Field's metal in its structure and, thus, alter its shape through the motion of its active components. 
In the soft state, the robot's main body can bend into circular arcs, which enables it to conform to surrounding curved objects. This variable geometry of the robot creates new motion modes which cannot be described by standard (i.e., fixed geometry) models. 
To this end, we develop a unified mathematical model that captures the differential kinematics of both rigid and soft states.
An optimised control strategy is further proposed to select the most appropriate phase states and motion modes needed to reach a target pose-shape configuration. 
The performance of our new method is validated with numerical simulations and experiments conducted on a prototype system.
The simulation source code is available at the \href{https://github.com/Louashka/2sr-agent-simulation.git}{GitHub repository}.
\end{abstract}

\begin{IEEEkeywords}
Hybrid soft-rigid robots, shape control, variable stiffness, adaptive morphology, mobile agents.
\end{IEEEkeywords}

% \fi
%\IEEEpeerreviewmaketitle

\section{Introduction}
\IEEEPARstart{A}{utonomous} mobile robots are one of the most prominent and well-established concepts in robotics. 
Self-contained design coupled with intelligent control systems allows mobile robots to perform a large variety of tasks. 
As such, they have been utilised in many industrial and service applications, including construction, conveyance, surface exploration, surveillance, entertainment, and others \cite{rubio_valero_llopis-albert_2019}. 

Despite their advantages, mobile robots face considerable limitations common to most conventional robots. 
Constant morphology and rigid structures undermine their ability to adapt to surroundings, handle delicate objects and safely interact with humans. 
Soft robotics can solve these issues \cite{kim_laschi_trimmer_2013}; however, it complicates (and possibly limits) the system's mobility and control. 
Existing soft robots operate via cumbersome actuators, lack high force capacity, and their flexible electronics are not developed enough to provide untethered design \cite{zhang_zhu_lin_jiao_zou_2020}. 

In this work, we aim to improve the performance of mobile robotic agents by leveraging paradigms of both conventional and soft robotics. 
We introduce a hybrid system with adaptive morphology that consists of a pair of wheeled locomotion units (LU's) connected by a variable-stiffness fibre (VSF). 
Sections with low-melting-point alloy (LMPA) within VSF serve as toggles between constant-shape and deformable modes. 
When the whole fibre is rigid, the robot acts as a conventional mobile robot. 
When LMPA is molten, VSF sections become soft; thus, a robot can reshape itself and conform to objects with various curved geometries \cite{shan_lu_majidi_2013, dna_tro_2018}. 
After the solidification of the fibre, the structure (which is now rigid) maintains a modified shape. 
LU's carry power and electronics, which provides autonomy to a mobile system. 
Omni-wheels at each unit enable the robot to undergo versatile deformations, as well as to move freely in a rigid state at any VSF configuration.    

This new class of hybrid systems with adaptive morphology is primarily inspired by nature \cite{pfeifer_lungarella_iida_2007}. 
Biological species exploit similar variable shape principles to gain additional functionalities and facilitate locomotion in dynamic environments \cite{mintchev_floreano_2016}. 
These benefits promoted research efforts in developing a new class of highly reconfigurable systems, including mobile robots \cite{quinn_nelson_2003, saranli_buehler_koditschek_2001, liu_edwards_althoefer_zhang_godaba_2022}. 
For example, Kim et al. applied variable morphology to extend the locomotion capabilities of Whegs \cite{kim_jung_kim_cho_chu_2014}, which enables a robot to traverse uneven surfaces. On flat surfaces, they are transformed into round wheels to facilitate high-speed motion. Similar idea lies behind robots that fold their whole body into the ball to roll on a plane \cite{xu_wang_fu_kang_2018, shah_powers_tilton_kriegman_bongard_kramer-bottiglio_2020}.
Another solution was demonstrated by Nygaard et al. \cite{nygaard_martin_torresen_glette_howard_2021}, in which a quadruped robot changed its legs length on the fly; A terrain-adaption algorithm allows this system to adjust its configuration to optimise energy efficiency and fit into unstructured environments.  

Several works have exploited variable morphology to create multi-functional devices such as self-reconfigurable robots \cite{yim_shen_salemi_2007}. However, most of these systems utilise rigid modules only. 
Many researchers have also worked towards self-foldable programmable matter with soft modules \cite{germann_maesani_pericet-camara_floreano_2014,hawkes_an_benbernou_tanaka_kim_demaine_rus_wood_2010}, but they typically lack mobility. 
Karimi et al. \cite{karimi_alizadehyazdi_jaeger_spenko_2021} presented one of the pioneer studies in hybrid self-reconfigurable systems. They developed a loop-shaped robot composed of rigid mobile units linked by a flexible membrane that becomes stiff by jamming its particles with an untethered vacuum system. 
This solution demonstrated promising results in locomotion and object handling, yet, its fixed number of active units may limit the system's scalability. 

This paper presents a new class of robots that can be used in numerous applications as a single robot or as a unit of a self-reconfigurable modular system and it will be called in a sequel a "Self-Reconfigurable Soft-Rigid Agent" (2SRA). 
Self-reconfigurable refers to the capability of our robot to reconfigure its shape and its potential to be a part of a self-reconfigurable modular system. 
The original contribution of this work is twofold: (i) The development and experimental validation of a novel mobile robotic system with phase transition capabilities; (ii) The derivation of a mathematical model that captures the hybrid soft-rigid behaviour of the system and its locomotion properties.

The rest of this paper is organised as follows: Sec. \ref{sec:materials} describes the robot's design, Sec. \ref{sec:models} derives the mathematical models, Sec. \ref{sec: controller} proposes the control strategy, Sec. \ref{sec:results} presents the experiments, and Sec. \ref{sec:conclusion} gives final conclusions.

\section{Materials and Methods}\label{sec:materials}

%A 2SR robot consists of two parts, which serve different goals and involve different fabrication and actuation techniques.

\subsection{Variable-Stiffness Fibre}

Variable-stiffness fibre (VSF) aims to alter the configuration and degrees of freedom of a 2SRA. For omnidirectional motion, the fibre must be stiff and capable of maintaining its shape. In the soft state, it should easily deform under the bending torque produced by the locomotion units. Therefore, the compliant features of the VSF are designed to have high flexibility and low elasticity. 

We adopted a method presented by Tonazzini et al.\cite{tonazzini_2016} to implement stiffness variability in the fibre. Desired mechanical features are achieved by integrating a low melting point alloy (LMPA) and a Joule heater for melting the alloy into the soft tube. Due to relatively low melting temperatures (47–62 $^{\circ}$C) and large solid/liquid phase change, LMPAs became popular in applications where controllable stiffness is required \cite{van_meerbeek_2016}. Among LMPAs, we chose Field's metal (composition by weight: 32.5\% bismuth, 51\% indium, and 16.5\% tin). It is characterised by high absolute stiffness when solid (Young modulus is more than 3GPa), a melting point around 62 $^{\circ}$C and low viscosity\cite{johann_2013,lipchitz_imbert_harvel_2013}. Unlike several other alloys, Field's metal does not contain toxic elements; therefore, it is easy and safe to utilise.

Figures \ref{fig: 2SR-design}(a, b) show the VSF and its inner structure. The fibre is designed as a serial chain of Field's metal VSS (short for the variable-stiffness segment) with plastic links on both ends. This approach decreases fabrication complexity and increases robustness since separately fabricated modular segments can be easily replaced if they fail. For experiments, we fabricated a fibre with three links and two VSS. Each plastic tube and VS segment are 30 mm and 40 mm long, respectively. Heaters are fabricated by coiling an enamelled copper wire (0.1 mm in diameter) around the silicone rod (5 mm in diameter). A silicone tube (internal diameter is 7 mm; external diameter is 8 mm) coaxial to the inner core is used to encapsulate the whole device. The Field's metal is distributed between the tube and the rod. NTC thermistors (MF52AT) embedded in both fibre segments measure the metal's temperature and provide feedback for stiffness control. Plastic links closing the ends of the device prevent metal leakage when it is in the liquid state and protect wires from external impact. 

At room temperature, the metal layer and the entire fibre are rigid. A soft state of the fibre is achieved by applying current (0.7-0.9 A) to the heater and melting the alloy. Separated heaters enable fibre segments to undergo phase transition independently, which provides greater deformation diversity.

\subsection{Locomotion Units}

We considered two main requirements when developing locomotion units:
\begin{enumerate}[label=\alph*)]
    \item In a soft state, LU's must be able to actuate each end of the variable-stiffness fibre and bend segments into arcs with the desired curvature. 
    \item In a rigid state, LU's must be able to maintain a 3 DoF motion of a 2SR agent on a plane with a given VSF configuration. 
\end{enumerate}

\begin{figure}[t]
    \centering
    \includegraphics[width=\columnwidth]{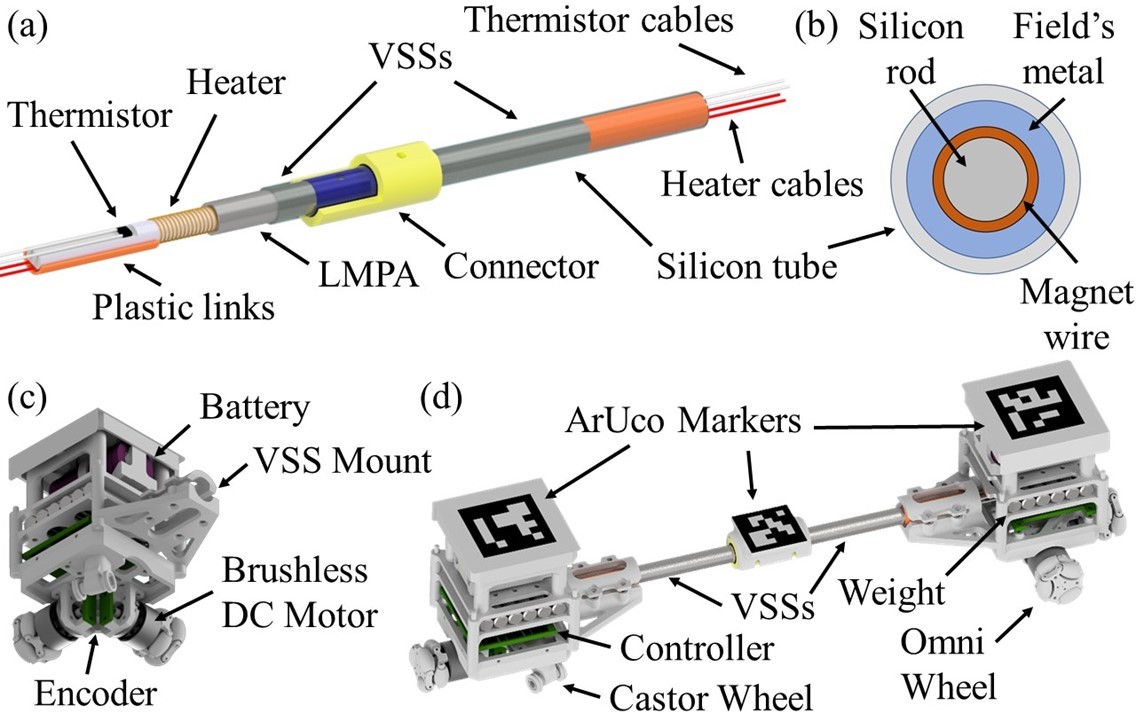}
    \caption{Variable-stiffness fibre (VSF) with two segments: (a) Perspective view; (b) Cross-sectional view. (c) Locomotion unit (LU). (b) Assembled 2SRA.}
    \label{fig: 2SR-design}
\end{figure}

To meet these conditions, LU's are designed as a nonholonomic robot (Fig.\ref{fig: 2SR-design}(c)) with two omni-wheels (radius of 10 mm) placed perpendicular to each other on the adjacent unit sides. Their rotation in the same direction induces the unit's rotation, while the counter rotation promotes the linear motion along the diagonal between the wheels. A castor wheel in the opposite corner from the omni-wheels is used to stabilise the unit. Steel bars are placed inside the LU frame to increase the unit weight, which helps to improve the wheels' contact with the surface and to resist the restoring force produced by deformed VS fibre. When VSF is rigid, units and the bridge create a single agent with four omni-wheels, see Fig.\ref{fig: 2SR-design}(d). Their mutual displacement guarantees that the 2SR robot stays omnidirectional regardless of the fibre shape.

A custom controller board with STM32F446RET6 microcontroller is designed to control brushless DC motors driving omni-wheels and actuate VSSs. AS5600 magnetic encoders increase the accuracy of motors' speed control. PWM output with analogue signal captured from the thermistor as feedback is used to control heaters. A wireless serial board provides communication with the server. All electronics are encapsulated in the 3d-printed plastic frame. ($46 \times 46 \times 78$ mm).

%Each moving block has two omni-wheels on neighbor sides, thus, they are oriented perpendicular to each other. A caster wheel is located in the far corner on the diagonal between the wheels to stabilize the module. When wheels rotate in the opposite directions the block moves either forward or backwards along the diagonal between them; when they rotate in the same direction the module undergoes both rotation and translation; otherwise, the actuation of one wheel produces a pure rotation. 

% The module has a square shape and its centre of mass is close to the geometric centre of the block; therefore, we assume that a rotation axis is also attached to the centre.

%Bla bla bla

\section{Modelling}\label{sec:models}

Integrating entirely different natures of rigid and soft bodies in one robot significantly expands its capabilities; however, it becomes challenging to derive a unified model. This section introduces a hybrid strategy that combines 2SRA configurations in both rigid and soft states and allows to switch between different motion modes in a simple manner. This model is adapted for a robot with two variable-stiffness segments, but it can be customised for any fibre design.  
% We start by converting wheel velocities into the set of robot velocities based on stiffness parameters of segments, use logarithmic spirals to approximate moving blocks trajectories and finish by determining a unified Jacobian matrix.

% \subsection{Notation}

% Throughout this paper, we denote column vectors and matrices by small and capital letters, respectively, e.g. $\mathbf v \in \mathbb{R}^h$ and $\mathbf M \in \mathbb{R}^{g \times h}$. A partial derivative of a vector-values function $\mathbf a = \mathbf{a(z)} \in \mathbb{R}^g$ with respect to $\mathbf z \in \mathbb{R}^h$ is represented as:
% \begin{equation*}
%     \partial_{\mathbf z}\mathbf a = \frac{\partial \mathbf a}{\partial \mathbf z} \in \mathbb{R}^{g \times h}.
% \end{equation*}

% We describe configuration of any mobile system via generalized coordinates $\mathbf q = [x, y, \theta]^\T$. Frames of the moving objects are denoted as $\{f\}$ and coordinates of $\{a\}$ frame with respect to $\{b\}$ frame are denoted as $\tensor[^{b}]{\mathbf q}{_{a}}$.

\subsection{Analysis of the System's Hybrid Kinematics}

Due to the 2SRA dual nature, we conduct kinematic analysis separately for rigid and soft states and then combine those pieces in one model. To simplify calculations, we assume fibre segments are inextensible and strictly bend into circular arcs with constant curvature. Also, we consider that motion of the units in the soft state is independent of each other. 

% Segments that are deformed simultaneously have a similar curvature. Different curvatures are achieved by gradual deformation when only one segment is soft and others are kept rigid. This allows to preserve different shapes of fibre arcs. 

\begin{figure}[t]
    \centering
    \includegraphics[width=\columnwidth]{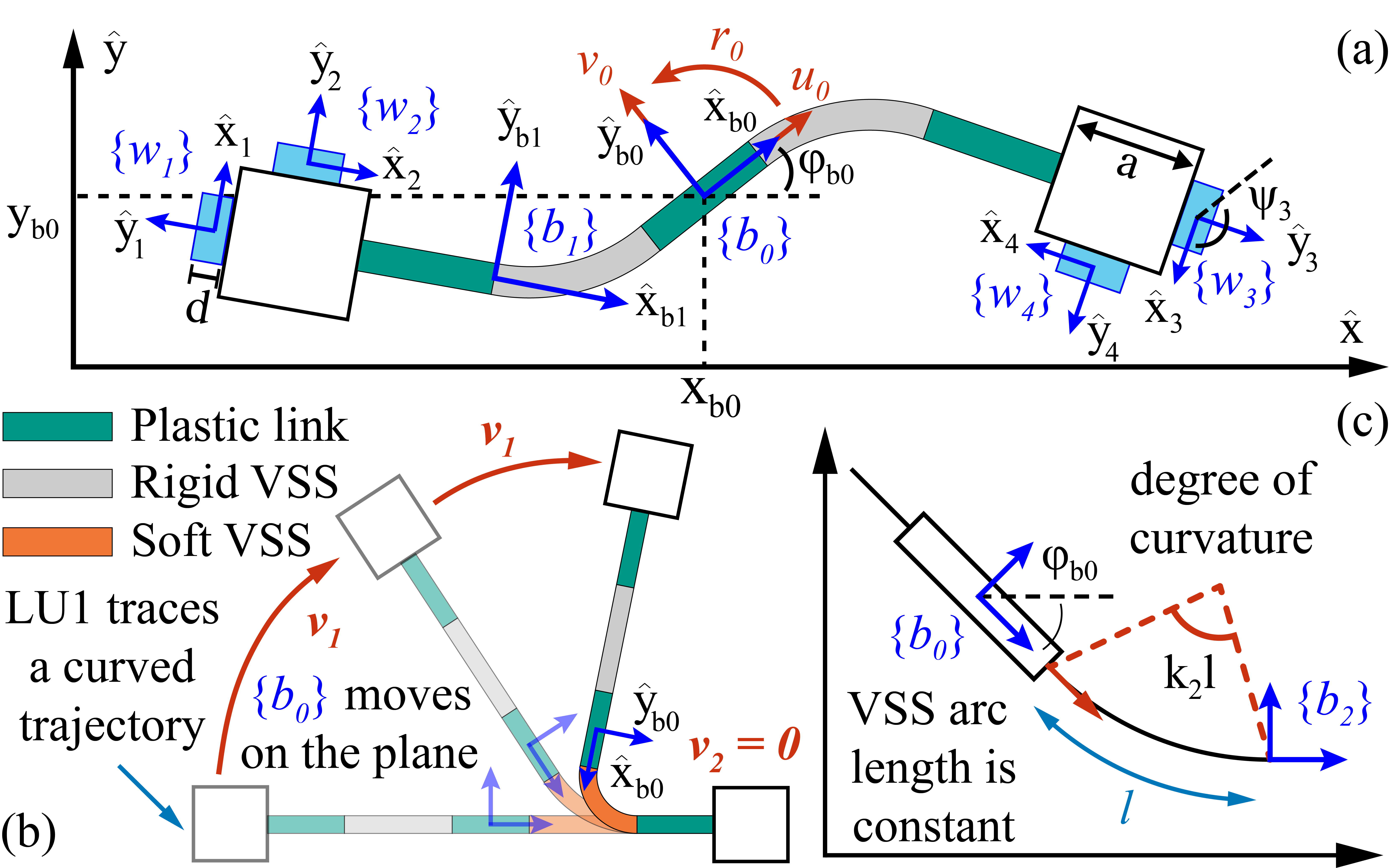}
    \caption{(a) Schematic representation of a 2SR robot with two variable-stiffness segments (VSS). (b) Reconfiguration of a 2SR robot when one of its segments is soft. The motion of a locomotion unit (LU) adjacent to the rigid segment results in a displacement of the body frame $\{b_{0i}\}$. (c) Geometrical features of constant curvature approximation for a VSS arc.}
    \label{fig: 2SR_schematics}
\end{figure}

Figure \ref{fig: 2SR_schematics}(b) demonstrates a scenario when one fibre segment is soft, and the 2SR agent can change its shape. Locomotion units can move together or one at a time. According to schematics in \ref{fig: 2SR_schematics}(a), omni-wheel indexing starts from the left side wheel of LU1 and follows clockwise. Robot's body frame $\{b_{0}\}$ is attached to the middle point of the middle plastic link. Two other frames $\{b_{1}\}$ and $\{b_{2}\}$ are attached to the VSS ends. The angles between these frames and the wheel frames $\{w_{i}\}$ $\left(i \in [1..4]\right)$ are $\beta_{1} = \frac{\pi}{2}$, $\beta_{2} = 0$, $\beta_{3} = -\frac{\pi}{2}$, and $\beta_{4} = \pi$ with the former two wheels associated with $\{b_{1}\}$ and the latter two wheels associated with $\{b_{2}\}$.

The biggest contribution to the fibre bending is provided by the side wheels with axes of rotation parallel to adjacent links. Velocities tangential to bending trajectory correspond to the maximum torque and thus are energy efficient. On the other hand, the motion of the other wheels promotes free rod deformation, which is challenging to perform since it requires high forces that are not feasible with the current design. Therefore, we neglect these wheels during VSS bending and keep their rotational speed at zero: $\omega_{1} = \frac{1}{\rho_{\omega}}v_{1}$, $\omega_{3} = -\frac{1}{\rho_{\omega}}v_{2}$, and $\omega_{2} = \omega_{4} = 0$, where $\omega_{i}$ is angular velocity of the $i^{th}$ wheel; $v_{1}$ and $v_{2}$ are velocities of the first and second units traversing some curved path; and $\rho_{\omega}$ is the wheel radius. In matrix representation, this relationship is defined as $\boldsymbol\omega = \mathbf{V}^{s} \boldsymbol{\upsilon}^{s}$:
\begin{equation}
    \mathbf{V}^{s} = \frac{1}{\rho_{\omega}}\begin{bmatrix}
                                            1 & 0 & 0 & 0 \\
                                            0 & 0 & -1 & 0 \\
                                       \end{bmatrix}^\T,
\label{eq: wheel_drive}
\end{equation}
where $\mathbf{V}^{s}$ is the wheels configuration matrix when 2SR agent is in a soft state; $\boldsymbol\omega = [\omega_{1}...\omega_{4}]^\T$ and $\boldsymbol{\upsilon}^{s} = [v_{1}, v_{2}]^\T$.

In the second scenario, the fibre is rigid, and 2SRA behaves like a conventional mobile robot. A remarkable feature of this state is that omnidirectional motion can be achieved at any fibre configuration since the relative position of omni-wheels ensures three degrees of freedom. For omnidirectional robots with four wheels, a generalised wheel model is defined as:
\begin{equation}
    \omega_{i} = \begin{bmatrix}
                              \frac{1}{\rho_{i}} & \frac{t\gamma_{i}}{\rho_{i}}
                              \end{bmatrix}
                            %   \begin{bmatrix}
                            %   cos\psi_{i} & \sin\psi_{i} & x_{C_{i}}\sin\psi_{i} - y_{C_{i}}\cos\psi_{i} \\
                            %   -\sin\psi_{i} & \cos\psi_{i} & x_{C_{i}}\cos\psi_{i} + y_{C_{i}}\sin\psi_{i}
                            %   \end{bmatrix}
                              \begin{bmatrix}
                              \text{c}\psi_{i} & \text{s}\psi_{i} \\
                              -\text{s}\psi_{i} & \text{c}\psi_{i}
                              \end{bmatrix}
                              \begin{bmatrix}
                              1 & 0 & -\tensor[^{b_{0}}]{y}{_{w_{i}}} \\
                              0 & 1 & \tensor[^{b_{0}}]{x}{_{w_{i}}}
                              \end{bmatrix}
                              \begin{bmatrix}
                              u_{0} \\
                              v_{0} \\
                              r_{0}
                              \end{bmatrix},
\label{eq: wheel-speed-gen}
\end{equation}
where $\omega_{i}$ is the $i^{th}$ wheel driving speed; $\rho_{i}$ is the radius of the $i^{th}$ wheel; $\gamma_{i}$ is an angle between the roller axis and the $\hat{x}_{i}$ axis of the wheel frame $\{w_{i}\}$; $\psi_{i}$ is an angle between the body frame $\{b_{0}\}$ and $\{w_{i}\}$; $\tensor[^{b_{0}}]{x}{_{w_{i}}}$ and $\tensor[^{b_{0}}]{y}{_{w_{i}}}$ are position coordinates of $\{w_{i}\}$ with reference to $\{b_{0}\}$; $u_{0}$, $v_{0}$, and $r_{0}$ are forward, lateral, and angular velocities of the robot. By $\text{s}\psi_{i}$ and $\text{c}\psi_{i}$ we denote sine and cosine of the angle $\psi_{i}$. Since all wheels have the same radius $\rho_{\omega}$, and $\gamma_{i}$ for omni wheels is zero, equation \eqref{eq: wheel-speed-gen} can be rewritten in the form of $\boldsymbol\omega = \mathbf{V}^{r} \boldsymbol{\upsilon}^{r}$:
\begin{equation}
    \mathbf{V}^{r}_i =  \begin{bmatrix}
                                \text{c}\psi_{i} & \text{s}\psi_{i} & \tensor[^{b_{0}}]{x}{_{w_{i}}}\text{s}\psi_{i} - \tensor[^{b_{0}}]{y}{_{w_{i}}}\text{c}\psi_{i}
                             \end{bmatrix}/\rho_{\omega},
\label{eq: wheel-speed-simple}
\end{equation}
where $\mathbf{V}^{r}_i$ is the $i^{th}$ row of a configuration matrix $\mathbf{V}^{r}$ when a 2SR agent is in a rigid state and $\boldsymbol{\upsilon}^{r} = [u_{0}, v_{0}, r_{0}]^\T$. Variables $\psi_{i}$, $\tensor[^{b_{0}}]{x}{_{w_{i}}}$ and $\tensor[^{b_{0}}]{y}{_{w_{i}}}$ depend on the fibre shape and thus to be recalculated accordingly. Analysis of the relationship between them and segments curvatures are given in the next section.

Combination of both approaches in $\boldsymbol\omega = \mathbf{V} \boldsymbol\upsilon$ gives a vector of input velocities $\boldsymbol\upsilon = (\boldsymbol{\upsilon}^{s}, \boldsymbol{\upsilon}^{r})$, in which either $v_{1}$ and $v_{2}$ or $u_{0}$, $v_{0}$ and $r_{0}$ are non-zero. This condition depends on the segments' stiffness. To represent the stiffness of the $j^{th}$ VSS, we introduce a boolean variable $\varsigma_{j}$, which is equal to zero when the segment is rigid and one when it is soft. 
Thus, we obtain a unified wheels configuration matrix:
\begin{equation}
    \mathbf{V}_{i} = 
    \begin{bmatrix}
                                    \sigma(i = 1) &
                                    -\sigma(i = 3) &
                                    \overline{\sigma} \text{c}\psi_{i} &
                                    \overline{\sigma} \text{s}\psi_{i} &
                                    \overline{\sigma} \tau_{i} 
                                \end{bmatrix}/\rho_{\omega},
\label{eq: config_matrix}
\end{equation}
where $\sigma = \varsigma_{1} \vee \varsigma_{2}$ and $\tau_{i}  = \tensor[^{b_{0}}]{x}{_{w_{i}}}\text{s}\psi_{i} - \tensor[^{b_{0}}]{y}{_{w_{i}}}\text{c}\psi_{i}$. Expression in parentheses is a conditional statement, $\vee$ is a logical disjunction, and notation $\overline{A}$ is a logical negation.

\subsection{Wheels Configuration}

For the robots with fixed geometry, matrix $\mathbf V$ from \eqref{eq: config_matrix} is constant. However, a 2SRA is capable of changing its shape, which entails the variation of the wheels' coordinates relative to the body frame. Therefore, the issue to address is determination of the $i^{th}$ wheel coordinates $\tensor[^{b_{0}}]{\mathbf q}{_{w_{i}}} = [\tensor[^{b_{0}}]{x}{_{w_{i}}}, \tensor[^{b_{0}}]{y}{_{w_{i}}}]^\T$ and $\psi_{i}$ for any robot configuration. Coordinates of the wheels with reference to the associated fibre ends are always constant and can be expressed  as:
\begin{equation}
    \tensor[^{b}]{\mathbf Q}{_{w}} = \begin{bmatrix}
        -h_{1} & -h_{2} & h_{1} & h_{2} \\
        0 & h_{3} & 0 & -h_{3}
    \end{bmatrix},
\label{eq: q_bC}
\end{equation}
where $h_{1} = \frac{2l_{1}+2a+d}{2}$, $h_{2} = \frac{2l_{1}+a}{2}$, and $h_{3} = \frac{a+d}{2}$; $a$ is the block side length, $d$ is the wheel thickness, and $l_{1}$ is the length of the plastic links at the ends of the fibre. Positional coordinates w. r. t. the robot body frame $\{b_{0}\}$ are further derived through a homogeneous transformation
\begin{equation}
    \begin{bmatrix}
        \tensor*[^{b_{0}}]{\mathbf q}{^{\T}_{w_i}} &
        1
    \end{bmatrix}^\T = \tensor[^{b_{0}}]{\mathbf T}{_{b_{j}}}
    \begin{bmatrix}
        \tensor*[^{b_{j}}]{\mathbf q}{^{\T}_{w_i}} &
        1
    \end{bmatrix}^\T,
\label{eq: homo-transformation}
\end{equation}
where $\tensor[^{b_{j}}]{\mathbf q}{_{w_{i}}}$ is the $i^{th}$ column of $\tensor[^{b}]{\mathbf Q}{_{w}}$ and $\{b_{j}\}$ denotes the fibre end associated to the $i^{th}$ wheel. Thus, the solution to the problem is reduced to mapping frames $\{b_{j}\}$ to the frame $\{b_{0}\}$, which depends on the VSS arc curve.

One of the approaches to describe the curve and position of the points along its length was suggested by Marchese et al.\cite{marchese_2014}. A soft arm is described as a sequence of circular arcs of signed curvature $\kappa$. With given starting orientation $\theta_{0}$, the position of any point along the arm is expressed as:
\begin{equation}
    \begin{gathered}
        x_{j}(p) = x_{j-1}\left(L_{j-1}\right) + \int_{0}^{p}\text{c}\left(\kappa_{j}p'+\theta_{0}\right)dp' \\
        y_{j}(p) = y_{j-1}\left(L_{j-1}\right) + \int_{0}^{p}\text{s}\left(\kappa_{j}p'+\theta_{0}\right)dp',
    \end{gathered}
\label{eq: marchese}
\end{equation}
where $j$ denotes the $j^{th}$ arm segment; $L_{j}$ is the length of the $j^{th}$ segment; $p$ is the position along the segment. Since all VS segments are inextensible, and their length $l$ remains constant during robot deformation, we derive a transformation matrix from this model as follows:
\begin{equation}
    \tensor[^{b_{0}}]{\mathbf T}{_{b_{j}}} = \begin{bmatrix}
                                \text{c}\alpha_{j} & \pm \text{s}\alpha_{j} & \mp \left(\frac{l_{0}}{2}+\frac{\text{s}\alpha_{j}}{\kappa_{j}}\right) \\
                                \mp \text{s}\alpha_{j} & \text{c}\alpha_{j} & \frac{1-\text{c}\alpha_{j}}{\kappa_{j}} \\
                                0 & 0 & 1
                            \end{bmatrix},
\label{eq: transform-matrix}
\end{equation}
where $l_{0}$ is the length of the middle link, $j$ is the segment index, and $\alpha_{j} = \kappa_{j}l$ is the degree of curvature, see Fig. \ref{fig: 2SR_schematics}(c). First sign in $\pm$ or $\mp$ corresponds to $j=1$ and the second one to $j=2$. Accordingly, we find an orientation of the wheel frame:
\begin{equation}
    \psi_{i} = \phi_{b_0} \mp \alpha_{j} + \beta_{i},
\end{equation}
where $\phi_{b_{0}}$ is orientation of the body frame $\{b_{0}\}$ w.r.t. to the global frame.

\subsection{LU Paths Estimation}

\begin{figure}[t]
    \centering
    \includegraphics[width=\columnwidth]{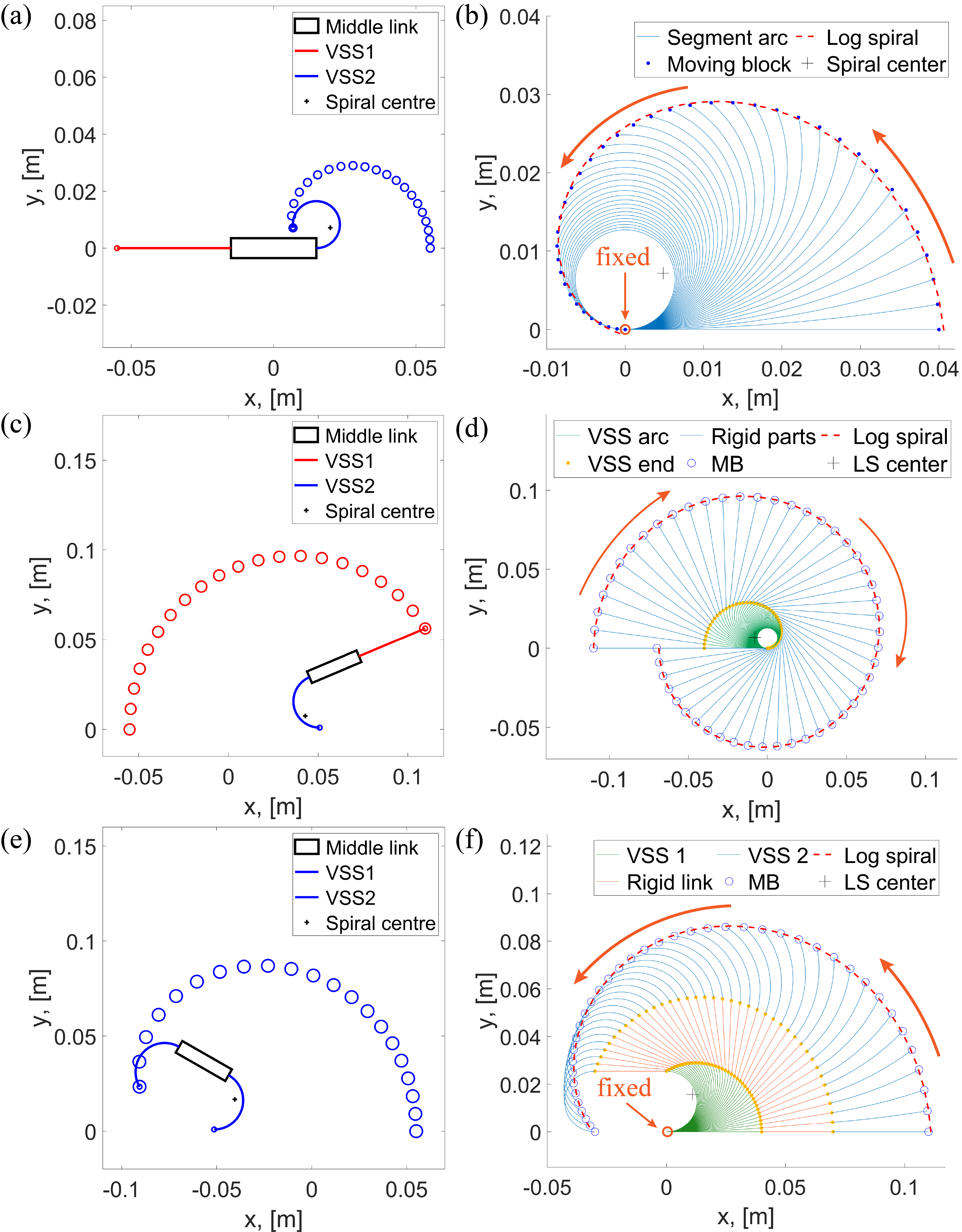}
    \caption{Shape deformation of a 2SR robot and corresponding logarithmic spirals when: (a), (b) one soft VSS is bent by an adjacent LU; (c), (d) one soft VSS is bent by the opposite LU; (e), (f) both segments are soft.}
    \label{fig: Jacobian_log_spiral}
\end{figure}

The further analysis involves estimating paths traced by LU's when either of the VS segments is soft. We consider segment deformation as an arc of a constant length and curvature. Assuming that VSS can bend into a full circle (though it is not feasible with a real robot), we can see that its tip traces a curve, which reminds a piece of a logarithmic spiral, see Fig. \ref{fig: Jacobian_log_spiral}. Matlab was used to fit the spiral to the segment arc points according to the following curve representation:
\begin{equation}
    \rho = a e^{b\theta}, \quad
    x = \rho\text{c}\theta, \quad
    y = \rho\text{s}\theta,
\label{eq: log_spiral}
\end{equation}
where $\theta$ is the spiral angle from $x$-axis; $\rho$ is the spiral radius at an angle $\theta$; and $a$ and $b$ are arbitrary constants. 

We distinguish three general motion modes at a soft state: one segment is soft, and the adjacent unit is moving; one segment is soft, and the opposite unit is moving; both segments are soft. Graphs in Fig. \ref{fig: Jacobian_log_spiral} demonstrate that each mode corresponds to a differently shaped logarithmic spiral; therefore, we first analyze each spiral and then combine their features into a unified kinematic model. 

\begin{table}[b]
\caption{\label{tab: spiral-param}Parameters of logarithmic spirals}
    \begin{tabularx}{\columnwidth}{|i|i|i|i|i|}
     \hline
     Spiral & $a/l$ & $b$  & $c_x/l$ & $c_y/l$ \\ 
     \hline
     I & 2.325 & $\pm 0.3165$  & -0.1223 & 0.1782 \\ 
     II & 3.3041 & $\pm  0.083$  & 0.1988 & 0.1640 \\ 
     III & 2.4471 & $\pm 0.2229$  & -0.2722 & 0.3949 \\ 
     \hline
    \end{tabularx}
\end{table}

The first considered case ($\varsigma_{1} = 0$, $\varsigma_{2} = 1$, $v_{1} = 0$, and  $v_{2} \neq 0$) is illustrated in Fig. \ref{fig: Jacobian_log_spiral}(a, b). By varying the fibre length $l$ from 1 to 100 cm in Matlab we determined that spiral parameter $b$, relationship between $l$ and $a$, and relationship between $l$ and spiral centre $\{c\}$ are constant. Parameters of all three spirals are listed in Table \ref{tab: spiral-param}. The sign of $b$ depends on curvature $\kappa$; if it is positive, then $b$ is negative and vice versa. In a 2SRA, the change of curvature $\kappa$ linearly depends on the change of $\theta$; therefore, we approximate their relationship by mapping their ranges $-\frac{\pi}{3} \leq \theta_{1} \leq \frac{7\pi}{3}$ and $-\frac{2\pi}{l} \leq \kappa_{2} \leq \frac{2\pi}{l}$:
\begin{equation}
    \kappa_{j} = 3\left(\theta_{1} - \pi\right)/(2l).
\label{eq: kappa_fit1}
\end{equation}

Motion of the unit adjacent to a rigid VSS ($\varsigma_{1} = 0$, $\varsigma_{2} = 1$, $v_{1} \neq 0$, and $v_{2} = 0$) resembles a logarithmic spiral even more clearly, see Fig. \ref{fig: Jacobian_log_spiral}(c, d). Here the absolute value of $b$ is sensitive to the ratio of the lengths of soft and rigid items. We applied fibre custom values used during fabrication. Relationship between spiral angle and segment curvature with given ranges $-\pi \leq \theta_{2} \leq 3\pi$ and $-\frac{2\pi}{l} \leq \kappa_{2} \leq \frac{2\pi}{l}$ is:
\begin{equation}
    \kappa_{j} = \left(\theta_{2} - \pi\right)/l.
\label{eq: kappa_fit2}
\end{equation}

The last case involves a 2SRA deformation with both segments being soft ($\varsigma_{1} = 1$, $\varsigma_{2} = 1$, $v_{1} = 0$, and $v_{2} \neq 0$), see Fig. \ref{fig: Jacobian_log_spiral}(e). Bending spreads uniformly along the fibre; thus, segment curvatures change equally. According to the corresponding logarithmic spiral (Fig. \ref{fig: Jacobian_log_spiral}(f)), units meet each other faster since segments' curvatures are summed up. Therefore, we limit each VSS to bend into a half-circle so that $-\frac{\pi}{3} \leq \theta_{3} \leq \frac{7\pi}{3}$ and $-\frac{\pi}{l} \leq \kappa \leq \frac{\pi}{l}$:
\begin{equation}
    \kappa_{j} = 3\left(\theta_{3} - {\pi}\right)/(4l).
\label{eq: kappa_fit3}
\end{equation}

\subsection{Unified Jacobian Matrix}

Configuration of a 2SR agent is fully defined by five generalised coordinates $\mathbf q = [x_{b_{0}}, y_{b_{0}}, \phi_{b_{0}}, \kappa_{1}, \kappa_{2}]^\T$, where $x_{b_{0}}$, $y_{b_{0}}$ and $\phi_{b_{0}}$ are position coordinates and orientation of the body frame $\{b_{0}\}$ w.r.t. to the global frame, and $\kappa_{j}$ is the curvature of the $j^{th}$ VSS. Then, the forward kinematics of a 2SR agent is described as:
\begin{equation}
    \mathbf{\dot{q}} = \mathbf J \boldsymbol\upsilon,
\end{equation}
where $\boldsymbol\upsilon = \mathbf{V^{\dagger}} \boldsymbol\omega$ and $\mathbf J$ is a Jacobian matrix. $\mathbf{V^{\dagger}}$ is the Moore-Penrose pseudo inverse of a configuration matrix $\mathbf V$ from \eqref{eq: config_matrix}.

In order to find a unified Jacobian matrix, we apply a state-centric approach as in the previous sections. We divide a Jacobian matrix into two parts: $\mathbf J = [\mathbf{J}^s, \mathbf{J}^r]\in\mathbb R^{5\times 5}$, where $\mathbf{J}^s$ and $\mathbf{J}^r$ are Jacobian matrices that correspond to the soft and rigid states, respectively. $\mathbf{J}^r\in\mathbb R^{5\times 3}$ consists of a rotation matrix $R_{z}(\phi_{b_{0}})$ and two zero rows. The matrix $\mathbf{J}^s\in\mathbb R^{5\times 2}$ maps velocity input commands into $\mathbf q$ coordinates according to the logarithmic spirals traced by locomotion units. Spirals are assigned serial numbers according to the order described in the previous section.
% Their motion can result either in only curvature change or also in a change of the body frame $\{b_{0}\}$ position and orientation. 

VSS curvature can be determined through the spiral angle $\theta$, which rate of change is proportional to the speed of the object tracing the spiral: $\dot{\theta} = \frac{1}{\rho\left(\theta\right)}v$. By substituting the angle $\theta_k$ by the curvature equations from \eqref{eq: kappa_fit1}-\eqref{eq: kappa_fit3}, we receive a relationship between curvature change and the LU velocity $v$:
\begin{equation}
    \begin{gathered}
        \dot{\kappa}_{j} = \frac{m_{k}}{l\rho_{k}(\kappa_{j})}v,
    \end{gathered}
\label{eq: kappa}
\end{equation}
where $k = [1..3]$ denotes the serial number of the spiral, $\mathbf{m} = \left[\frac{3}{2}, 1, \frac{3}{4}\right]^\T$ is the set of constants that correspond to the $k^{th}$ spiral, and $\rho_{k}(\kappa_{j}) = a_{k}\text{exp}\left({b_{k}(\frac{\alpha_{j}}{m_{k}}+\pi)}\right)$. Further, we denote the proportionality coefficient as $K_{k,j} = \frac{m_{k}}{l\rho_{k}(\kappa_{j})}$.

The next step is to find the body frame coordinates. In case I, the middle plastic link is stationary, thus $\dot{x}_{b_{0}} = \dot{y}_{b_{0}} = \dot{\phi}_{b_{0}} = 0$. States II and III involve the motion of the middle link with a body frame attached to it. It's orientation is equal to the orientation of the tangent at the VSS arc point closest to the body frame, which is calculated as $\phi_{b_{0}} = \phi_{b_{j}} \pm \alpha_{j}$, see Fig. \ref{fig: 2SR_schematics}(c). It implies that the rate of change of the robot orientation linearly depends on the rate of curvature change:
\begin{equation}
    \dot{\phi}_{b_{0}} = \frac{m_{k}}{\rho_{k}(\kappa_{j})}v.
\label{eq: jacobain_angle}
\end{equation}
We denote the proportionality coefficient as $\Phi_{k,j} = \frac{m_{j}}{\rho_{j}(\kappa)}$.

\begin{figure}[t]
    \centering
    \includegraphics[width=\columnwidth]{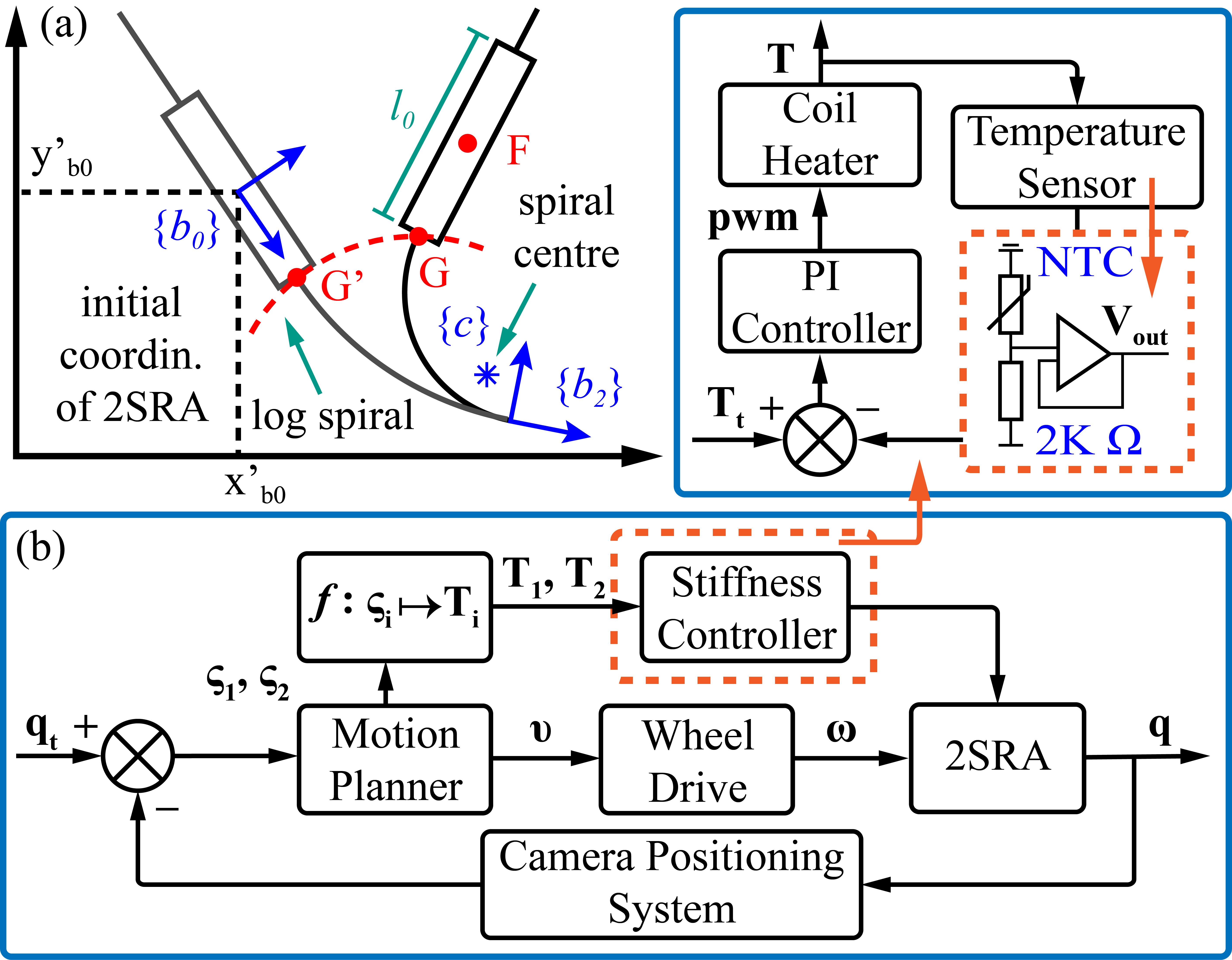}
    \caption{(a) The relationship between 2SR robot coordinates and the geometry of the logarithmic spiral I. (b) 2SRA control strategy.}
    \label{fig: Jacobian_geometry_control}
\end{figure}

Schematic representation of 2SRA geometry during shape deformation in Fig. \ref{fig: Jacobian_geometry_control}(a) demonstrates the relationship between fibre frames and logarithmic spiral I. The connection point \emph{G} between VSS and the middle link lies on the spiral and its position w.r.t. the spiral centre $\{c\}$ is found according to \eqref{eq: log_spiral}. Since the position of the point \emph{F} attached to the body frame origin is fixed relative to \emph{G}, we find:
\begin{equation}
    \tensor[^{c}]{\emph{F}}{} = \begin{bmatrix}
                                   \frac{l_{0}c\alpha_{j}}{2}+\rho_1(\kappa_{j})c\frac{\alpha_{j}}{m_{1}} &
                                  \frac{l_{0}s\alpha_{j}}{2}+\rho_1(\kappa_{j})s\frac{\alpha_{j}}{m_{1}}
                              \end{bmatrix}^\T.
\end{equation}
We can express frames $\{b_{j}\}$ through $\{b_{0}\}$ and $\{c\}$. Thus, position coordinates of the point \emph{F} with respect to the global frame $\{o\}$ can be expressed as a series of kinematic transformations as:
$\tensor[^{o}]{\emph{F}}{} = \tensor[^{o}]{\mathbf T}{_{b_{0}}}\tensor[^{b_{0}}]{\mathbf T}{_{b_{j}}}\tensor[^{b_{j}}]{\mathbf T}{_{c}}\tensor[^{c}]{\emph{F}}{}$, which partial derivatives with respect to $\kappa_{j}$ provides the relationship between the rate of change of the robot position coordinates, the LU's velocities $v_j$, and their logarithmic trajectories:
\begin{equation}
    x_{b_{0}} = \frac{\partial \tensor[^{o}]{\emph{F}}{_{x}}}{\partial \kappa_{j}}\frac{\partial\kappa_{j}}{\partial t}, \quad
    y_{b_{0}} = \frac{\partial \tensor[^{o}]{\emph{F}}{_{y}}}{\partial \kappa_{j}}\frac{\partial\kappa_{j}}{\partial t}.
    \label{eq: F_derivative}
\end{equation}

Denoting $\boldsymbol\Delta_{k,j} = \frac{\partial \tensor[^{o}]{\emph{F}}{}}{\partial \kappa_{j}}K_{k,j}$, $s_{1} = \overline{\varsigma_{1}}\varsigma_{2}$, $s_{2} = \varsigma_{1}\overline{\varsigma_{2}}$, $s_{3} = \varsigma_{1}\varsigma_{2}$, and $\mathbf{C}_{k_{1},j}^{k_{2},k_{3}} = [\boldsymbol\Delta_{k_{1},j}^\T, (-1)^{j+1}\Phi_{k_{1},j}, K_{k_{2},1}, K_{k_{3},2}]^\T$ we get:
\begin{equation}
    % \mathbf{J}^s = \begin{bmatrix}
    %                 \overline{\varsigma_{1}}\varsigma_{2}\Delta_{2,2}^x + \varsigma_{1}\varsigma_{2}\Delta_{3,2}^x & \varsigma_{1}\overline{\varsigma_{2}}\Delta_{2,1}^x + \varsigma_{1}\varsigma_{2}\Delta_{3,1}^x \\
    %                 \overline{\varsigma_{1}}\varsigma_{2}\Delta_{2,2}^y + \varsigma_{1}\varsigma_{2}\Delta_{3,2}^y & \varsigma_{1}\overline{\varsigma_{2}}\Delta_{2,1}^y + \varsigma_{1}\varsigma_{2}\Delta_{3,1}^y \\
    %                 -\overline{\varsigma_{1}}\varsigma_{2}\Phi_{2,2}-\varsigma_{1}\varsigma_{2}\Phi_{3,2} & \varsigma_{1}\overline{\varsigma_{2}}\Phi_{2,1}+\varsigma_{1}\varsigma_{2}\Phi_{3,1} \\
    %                 \varsigma_{1}\overline{\varsigma_{2}}K_{1,1} + \varsigma_{1}\varsigma_{2}K_{3,1} & \varsigma_{1}\overline{\varsigma_{2}}K_{2,1} + \varsigma_{1}\varsigma_{2}K_{3,1} \\
    %                 \overline{\varsigma_{1}}\varsigma_{2}K_{2,2} + \varsigma_{1}\varsigma_{2}K_{3,2} & \overline{\varsigma_{1}}\varsigma_{2}K_{1,2} + \varsigma_{1}\varsigma_{2}K_{3,2}
    %              \end{bmatrix}
    \mathbf{J}^s = \begin{bmatrix}
                       s_{1} & -s_{1} & s_{2} & s_{1} \\
                       s_{2} & s_{2} & s_{2} & s_{1}
                   \end{bmatrix}^\T \circ 
                   \begin{bmatrix}
                       \mathbf{C}_{2,2}^{1,2} \\
                       \mathbf{C}_{2,1}^{2,1}
                   \end{bmatrix}^\T + s_{3}
                   \begin{bmatrix}
                       \mathbf{C}_{3,2}^{3,3} \\
                       \mathbf{C}_{3,1}^{3,3}
                   \end{bmatrix}^\T,
\end{equation}
where $\circ$ denotes the element-wise matrix multiplication.

\section{Controller Design}\label{sec: controller}

A distinctive feature of a 2SR agent is the real-time transition between rigid and soft states, which extends locomotion capabilities and increases the robot's performance in conforming to the surroundings. This quality is exploited by a single-agent control strategy designed for scenarios with no obstacles between the initial and target configurations. A block diagram in Fig. \ref{fig: Jacobian_geometry_control}(b) shows that the strategy comprises three main components: Motion Planner, Wheel Drive, and Stiffness Controller.

% \begin{figure}[!b]
%     \centering
%     \includegraphics[width=\columnwidth]{Images/Control.png}
%     \caption{Design of a 2SR robot control strategy.}
%     \label{fig: 2SR_control}
% \end{figure}

At any given moment, a 2SR agent can switch between four locomotion modes: omnidirectional motion and motion along three logarithmic spirals. A Motion Planner aims to select a sequence of these modes most suitable for the given task. Possible configurations of the VSF are defined by a set $\boldsymbol{\varsigma} = [[0, 0], [0, 1], [1, 0], [1, 1]]^\T$. Each iteration of the trajectory formation calculates unified Jacobian matrices corresponding to the elements of $\boldsymbol{\varsigma}$. They are further supplied to the inverse kinematics equations. Among the Euclidean distances between the target configuration and the set of potential configurations, we choose the smallest one. However, in this approach, we might get solutions involving frequent changes in segments' stiffness. Phase transitions of the Field's metal take time; therefore, these situations are highly undesirable. We avoid this issue by maintaining established stiffness values as long as the robot drives towards the target. Once an intermediate configuration stops updating, the locomotion mode is changed to the appropriate option. The Motion Planner pseudocode is given in Algorithm \ref{alg: motion-planner}.

\begin{algorithm}[hb]
\DontPrintSemicolon
\caption{2SR Agent Motion Planner}
\label{alg: motion-planner}
    \KwInput{Initial configuration $q_0$, target configuration $q_t$}
    \KwOutput{List of stiffness values $\Omega$, list of velocities $\mathcal{V}$}
    $\varsigma \leftarrow $ set of possible VSF stiffness values \\
    $\Omega, \mathcal{V} \leftarrow$ empty lists \\
    \While{$||q \textbf{}- q_{t}|| > 0$} {
        \For{$i=1$ \KwTo $4$} {
            $J \leftarrow$ hybridJacobian$\left(q_{0}, q, \varsigma[i]\right)$ \\
            $\upsilon[i] \leftarrow J^{-1}\lambda (q_{t} - q)$ \tcp*{$\lambda$ - feedback gain} 
            $\dot{q} \leftarrow J\upsilon[i]$ \\
            $q^{*}[i] \leftarrow q + \dot{q}dt$
        }
        $\Delta q^* \leftarrow ||q^{*} - q_{t}||$ \\
        $i^{*} \leftarrow  \underset{i}{\mathrm{arg\,min}} \, \Delta q^*$  \\
        \If{$i^{*} \neq i_{prev}$} {
            \If{$||q - q^{*}[i_{prev}]|| > $ 0} {
                $i^{*} \leftarrow i_{prev}$ \\
            }
        }
        $q \leftarrow q^{*}[i^{*}]$ \\
        Add $\varsigma[i^{*}]$ to $\Omega$; Add $\upsilon[i^{*}]$ to $\mathcal{V}$ 
    }
\end{algorithm}

A sequence of 2SRA velocities returned by the Motion Planner is converted into the wheels' velocities by Wheel Drive defined by the model derived in \eqref{eq: wheel_drive}-\eqref{eq: marchese}. A sequence of VSSs' stiffness values is mapped to the temperature values that are required to either melt the Field's metal (taken with margin) for the soft state or to cool it down for the rigid state: $T_{j}(\varsigma_{j}) = 65^{\circ}$ if $\varsigma_{j} = 1$ else $25^{\circ}$.

The stiffness controller maintains the melting point when it is requested and notifies the main controller when the required temperature has been achieved. A PI controller regulates the current applied to the heater and receives feedback from the sensor. Temperature is measured via an NTC MF52AT thermistor. We constrain its operating range to $[0^{\circ}...85^{\circ}]$ to avoid overheating the copper coil. This range is mapped to voltage output $[1.1,\ldots,3.3]$ V. 

\section{Results}\label{sec:results}

\subsection{Experimental Setup and Simulation}  \label{sec: simulation}

\begin{figure}[t]
    \centering
    \includegraphics[width=\columnwidth]{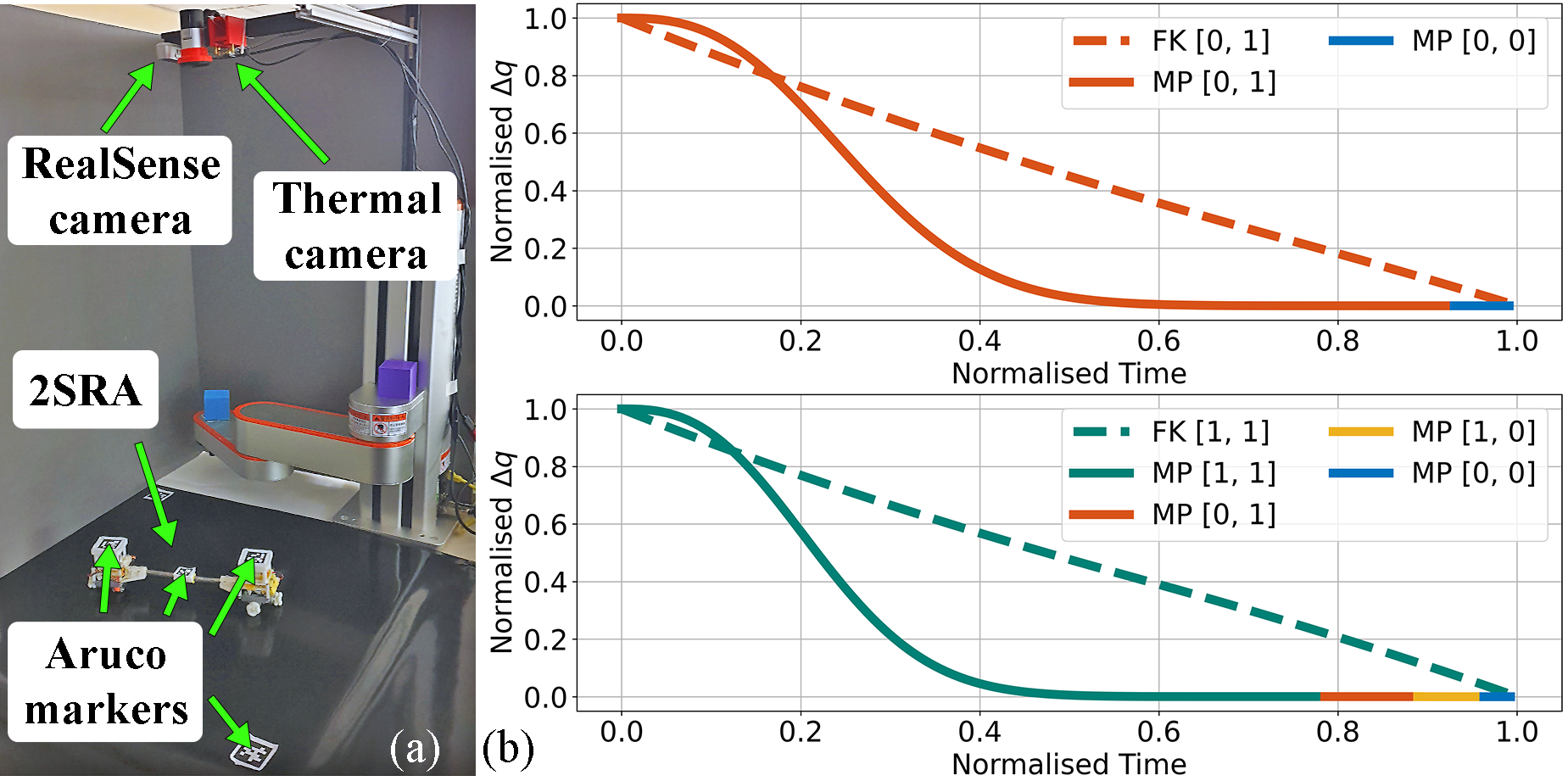}
    \caption{(a) Experimental setup. (b) Simulation results: difference between configuration trajectories generated by Forward Kinematics (FK) and Motion Planner (MP).}
    \label{fig: setup-sim}
\end{figure}

Figure \ref{fig: setup-sim}(a) shows an experimental set-up, which involved a 2SR agent with three Aruco markers attached to the locomotion units and the central VSF link, RealSence camera, thermal camera and a server. The coordinates of the VSF marker corresponded to the position coordinates of the robot $b_{0}$ frame, while the LU markers were used to estimate the segments' curvature with the assumption that it is constant along the arc. The agent's motion was kept within the tracked boundaries defined by Aruco markers, which, with the other markers, were captured by the RealSense camera. A thermal camera was used to analyse VSS temperature. 

Before conducting experiments, we passed our model through a simulator deployed in a Python environment. Plots in \ref{fig: setup-sim}(b) compare the forward kinematics (FK) and Motion Planner (MP) trajectories. We can see that the configuration change in FK is linear, while MP generates a slower start but reduces the Euclidean distance between initial and target configurations faster. It happens because the Algorithm \ref{alg: motion-planner} works similarly to the gradient descent, which results in a slight discrepancy between the positional coordinates of the target configuration and the configuration reached by the robot upon deforming into the desired shape. Therefore, 2SRA has to switch stiffness and motion modes to fit the target completely. Within 100 simulations, a robot exploited at most four motion modes, in which the last one was always rigid.

\begin{figure*}[t]
    \centering
    \includegraphics[width=\textwidth]{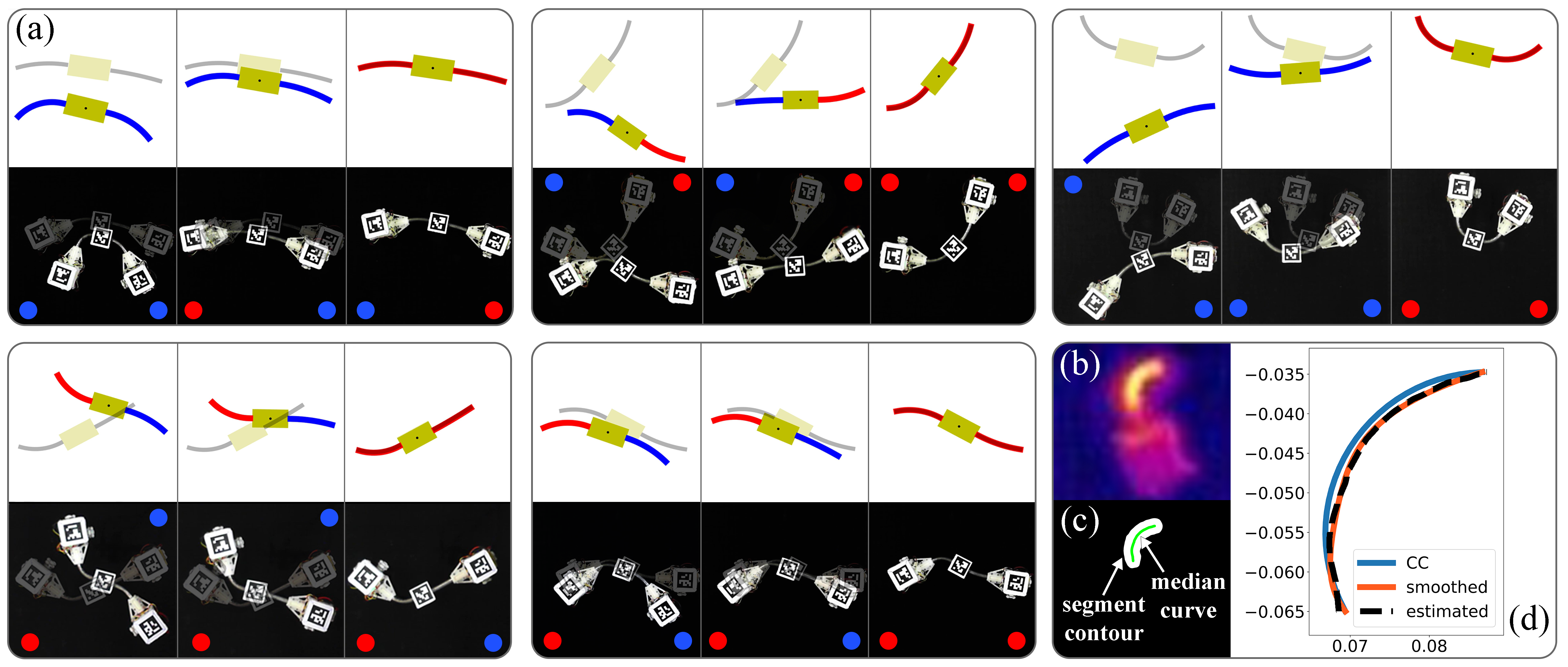}
    \caption{(a) Five different scenarios of the pose-shape formation when a robot has a task to reach some target configuration. The top row shows the behaviour of a 2SR agent in the simulator, and the bottom row is the robot's real-case motion/deformation. Blue colour corresponds to a soft segment; red colour denotes a rigid segment. A faded image shows a target, while the bright one is a configuration within a current trajectory. (b) Thermal image of a variable-stiffness segment in a soft state. The bight-coloured area shows the high-temperature zone corresponding to the molten LMPA inside the segment's shell. (c) Contours and a median curve of the bent segment extracted from a thermal image. (d) Comparison of the median curve, its smoothed version and the approximated curve with constant curvature along its length.}
    \label{fig: main-exp}
\end{figure*}

\subsection{Curvature Estimation}

The proposed hybrid model uses a constant curvature (CC) approximation of the fibre segments. We evaluated the actual curvature distribution along the segment to validate a CC assumption. One segment was maintained rigid and straight while its adjacent locomotion unit was fixed on the plane. Another segment was bent in both directions with a degree of curvature varying from $[-\pi ... \pi]$. The hot area of the soft VSS was extracted from the images captured by the thermal camera, see Fig. \ref{fig: main-exp}(b) and further processed for curvature analysis. Binary image in Fig. \ref{fig: main-exp}(c) shows the segment contours and median curve constructed via the OpenCV library. Figure \ref{fig: main-exp}(d) compares the estimated, smoothed and approximated segment arcs. Finally, calculating the gradient in each point of the smoothed median curve returned a corresponding array of curvatures. 
Test trials were repeated five times.

Results of the experiment presented in figure \ref{fig: main-exp-data}(d) show that the constant curvature and the mean curvature of the median curves extracted from the thermal images are quite similar at the small values, and the difference between them increases with larger bending. The overall mean square error is 41.32, and the average standard deviation is 12.93. According to the obtained results, CC approximation adequately represents the segment curvature.

\subsection{Pose-Shape Formation}

Pose-Shape Formation tasks involve reaching a given target with simultaneous motion on the plane and bending the segments. We implemented a control strategy provided in Section \ref{sec: controller} and conducted five experiments with different initial and target configurations. They were obtained by manually deforming the robot, placing it in the desired spot and measuring the corresponding configuration through a RealSense camera. The same inputs were tested in the simulator to assess the errors occurring in the real-case scenario. 
% There were five experimental trials in total.

% (c) Decrease of Euclidean distance $\Delta q$ by the 2SR robot while approaching the target. (d) Comparison of the constant curvature and curvature distribution extracted from the thermal imaging

\begin{figure}[H]
    \centering
    \includegraphics[width=\columnwidth]{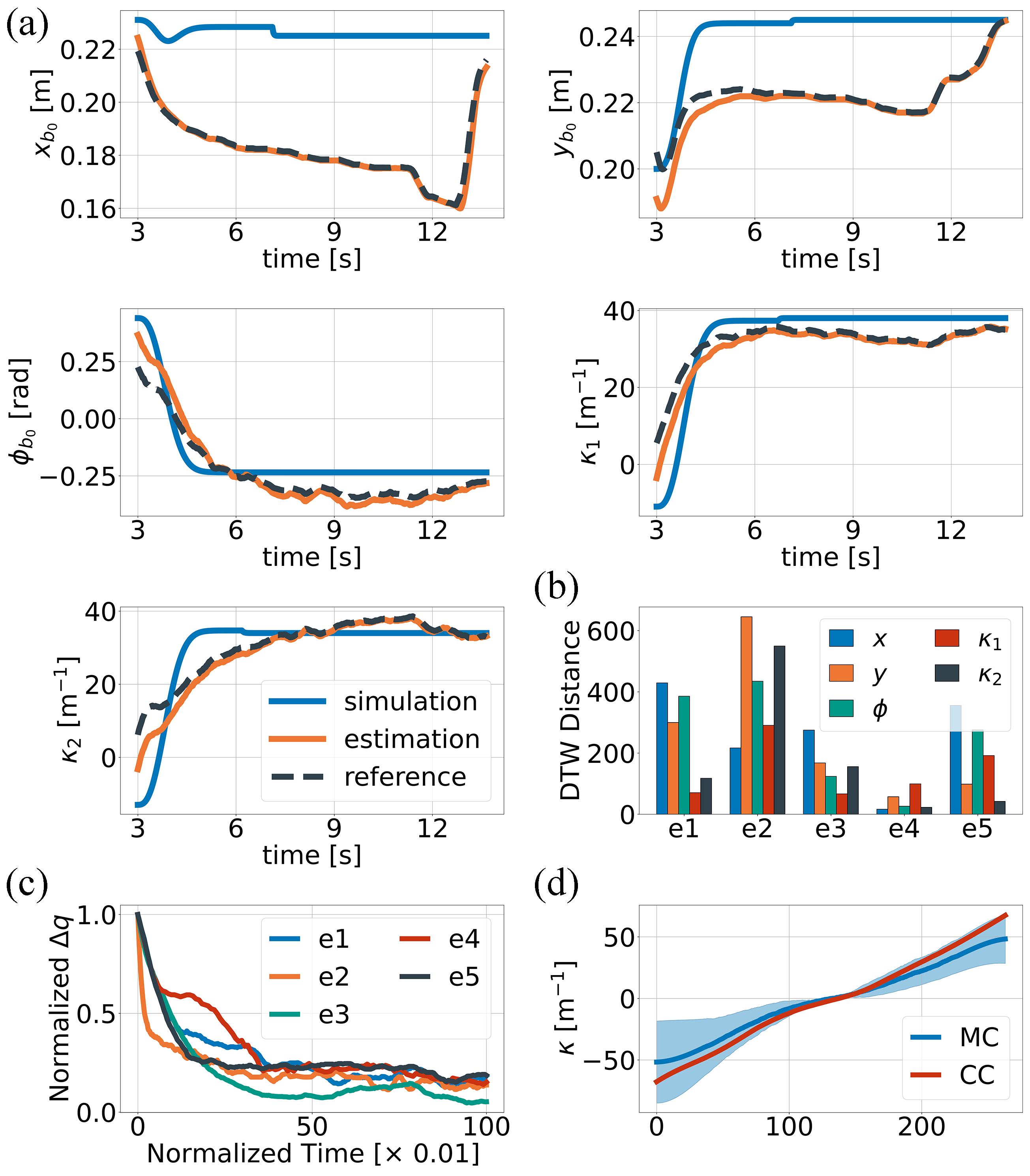}
    \caption{(a) Comparison of the configuration trajectories obtained during the Pose-Shape Formation experiment (3rd case) and the corresponding simulation. Reference is the configuration returned by a Motion Planner and further supplied to the robot; Estimated value is the configuration measured via camera and Aruco markers. (b) Quantified difference (DTW distance) between the configuration trajectories from the Pose-Shape Formation experiments and corresponding simulations. (c) Minimization of the error $\Delta q$ during task execution. (d) Comparison of the VSS constant curvature (CC) approximation and mean curvature (MC) of the segment arc extracted from thermal images. Solid coloured area demonstrates curvature distribution along the arc.}
    \label{fig: main-exp-data}
\end{figure}

All trials differ in VSF stiffness values, motion modes and execution time. Figure \ref{fig: main-exp}(a) demonstrates the execution process of the tasks and corresponding simulations. In both scenarios, a 2SRA shows a similar movement pattern. At each moment of the robot motion/deformation, we assessed a reference configuration selected by a Motion Planner and an estimated configuration measured through camera. The comparison of the trajectories obtained during the 3rd experiment and its simulation is provided in Fig. \ref{fig: main-exp-data}(a). The best fit is displayed for the curvature and orientation coordinates. As was mentioned in Sec. \ref{sec: simulation}, a discrepancy in the position trajectories occurs since the algorithm aims to minimize the Euclidean distance $\Delta q = ||q - q_t||$ in the fastest way, which preferences the curvature over position coordinates. Therefore, the robot fits $x_{b_{0}}$ and $y_{b_{0}}$ the last. 
Figure \ref{fig: main-exp-data}(c) shows how the robot approaches the target. We stopped the motion as soon as $\Delta q$ hit a given threshold instead of trying to reach zero, which helped avoiding frequent stiffness alternation.
Bar plot in Fig. \ref{fig: main-exp-data}(b) summarises the results of all experiment cases by calculating the Dynamic Time Warping (DTW) distance between the simulated and experimental configuration trajectories.
These errors are not significant, and a 2SR robot is capable of approaching the target under the proposed hybrid model. 

\subsection{Applications}

\begin{figure}[t]
    \centering
    \includegraphics[width=\columnwidth]{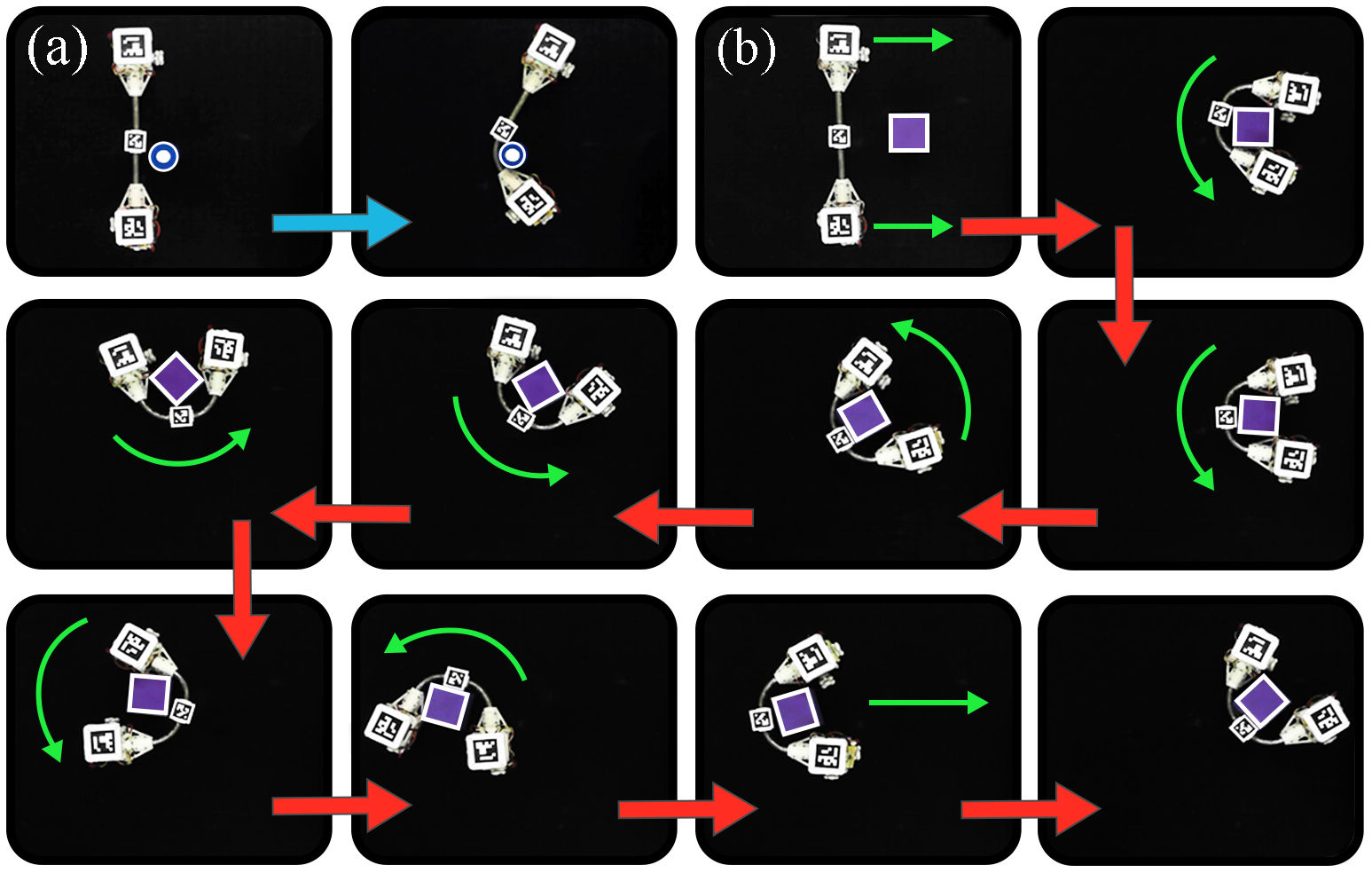}
    \caption{2SR agent performing objects handling tasks: (a) conforming to an object with a cylindrical shape. (b) gripping and moving a cube on the plane.}
    \label{fig: applications}
\end{figure}

The adaptive nature of a 2SR agent makes it applicable to a large variety of tasks. One of its main advantages is that a robot can conform to objects with different shapes. In Fig. \ref{fig: applications}(a), a 2SR agent approaches a cylindrical object and bends one of its segments until it fits the given shape. Another example shown in Fig. \ref{fig: applications}(b) demonstrates the object manipulation by a 2SRA. Execution of this task runs as follows: a robot approaches the target, makes both segments soft, grasps the cube and switches back to the rigid state. After fibre solidification, a robot is capable of moving freely on a plane while handling the cube. 

These are a few examples of how a 2SRA can be used. Potential applications are not limited to any specific area since, unlike other mobile robots, a 2SR agent aims to combine the best features of conventional and soft robots with adaptive morphology. Comparison of a 2SRA with the state of the art is given in Table \ref{tab: state-of-art}. 

\begin{table}[h]
\caption{\label{tab: state-of-art}Comparison of a 2SR agent with the state of art}
    \begin{tabularx}{\columnwidth}{|Z|s|s|s|s|}
     \hline
      & \cite{liu_edwards_althoefer_zhang_godaba_2022} & \cite{xu_wang_fu_kang_2018}  & \cite{karimi_alizadehyazdi_jaeger_spenko_2021} & 2SRA \\ 
     \hline
     Shape shifting & \cmark & \cmark  & \cmark & \cmark \\ 
     Conforming to surroundings & \cmark & \xmark  & \cmark & \cmark \\ 
     Variable stiffness & \xmark & \xmark  & \cmark & \cmark \\ 
     Variable motion modes & \cmark & \cmark  & \xmark & \cmark \\ 
     High-speed locomotion  & \cmark & \cmark  & \xmark & \cmark \\ 
     Untethered actuation & \xmark & \cmark & \cmark & \cmark \\
     \hline
    \end{tabularx}
\end{table}

\section{Conclusion}\label{sec:conclusion}

The paper presented a new type of mobile robot that can behave either as a conventional rigid mobile agent or as a hybrid soft robot. It consists of two locomotion units connected by a serial variable-stiffness bridge. VSF enables the transition between rigid and soft states and the robot morphing. LU's carry power and electronics, perform 2d motion and actuate the fibre. A unified hybrid model is developed to describe the behaviour scenarios of a 2SR agent in both states. Shape deformation and unit motion when some or all VSF segments are soft are described based on approximated logarithmic spirals traced by the units. The proposed design and model were tested in simulation and physical experiments.

Experiments in reaching the target and handling the objects demonstrated promising results. A 2SRA benefits from the ability to adapt to the surroundings and chooses the most appropriate stiffness and motion mode to execute the task. In future work, we will focus on reducing the phase transitioning time and developing a modular system that contains multiple 2SR agents. It will significantly increase their performance as a part of the swarm and expand the area of their applications. 

% \newpage 

\bibliographystyle{IEEEtran}
\bibliography{IEEEabrv.bib}


% \end{thebibliography}

\end{document}